# On the Algebra in Boole's *Laws of Thought*

Subhash Kak

**Abstract.** This article explores the ideas that went into George Boole's development of an algebra for logic in his book *The Laws of Thought*. The many theories that have been proposed to explain the origins of his algebra have ignored his wife Mary Boole's claim that he was deeply influenced by Indian logic. This paper investigates this claim and argues that Boole's focus was more than a framework for propositions and that he was trying to mathematize cognitions as is assumed in Indian logic and to achieve this he believed an algebraic approach was the most reasonable. By exploring parallels between his work and Indian logic, we are able to explain several peculiarities of his algebraic system.

**Introduction**
There is continuing interest in the antecedents to George Boole's *The Laws of Thought* [1] and an ongoing discussion on how he created a system in which the algebraic and logical calculi are not in perfect accord [2]. The sum and difference operations that Boole denotes by + and − are neither the standard set-theoretical union (between arbitrary sets) nor the set-theoretical difference.

The discrepancy between algebra and logic seen in Boole's system is problematic given that it was a period where these questions were much in discussion and his friend Augustus De Morgan (1806-1871) had also presented a formal framework for logic [3][4]. Boole (1815-1864) was the younger colleague of De Morgan, and the two of them carried on an extensive correspondence for years that was only published in 1982 [5]. However, this correspondence, which was only published in 1982, shows they ignored each other's work suggesting that they were still in the process of developing their ideas and they saw their work as somewhat tentative.

Another interesting perspective related to Boole's work is provided by his wife Mary Boole (1832-1916), who, during her times, was a well-known writer on mathematical subjects. She claims [6] that her husband as well as De Morgan and Charles Babbage were influenced deeply by Indian logic and her uncle George Everest (1790-1866), who lived for a long time in India and whose name was eventually given to the world's highest peak, was the intermediary of these ideas. She adds [6]: "Think what must have been the effect of the intense Hinduizing of three such men as Babbage, De Morgan, and George Boole on the mathematical atmosphere of 1830–65," further speculating that these ideas also influenced the development of vector analysis and modern mathematics. Although the statement of Mary Boole is well known, I know of no scholarly study that has attempted to explore the question of the Influence of Indian logic on Boole's work.

So let's follow the line of thought that arises from Mary Boole's claim. An account of Indian logic was presented by the Sanskritist H.T. Colebrooke at a public meeting of the Royal Asiatic Society 1824 and it was widely discussed in the scholarly world. Robert Blakely had a chapter on "Eastern and Indian Logic" in his 1851 book titled *Historical Sketch of Logic* in which he suggested that this knowledge has "been brought prominently forward among European *literati*" [7]. De Morgan admitted to the significance of Indian logic in his book published in 1860: "The two races which have founded the mathematics, those of the Sanscrit [an alternative early spelling of "Sanskrit"] and Greek languages, have been the two which have independently formed systems of logic" [8]. One must therefore accept the correctness of Mary Boole's statement that De Morgan, George Boole and Babbage were cognizant of Indian logic even though George Boole does not mention Indian logic texts or the larger Indian tradition in his book. We must ascribe this to the fact that while per Mary Boole's claim, George Boole and others knew of Indian logic, they were apparently not knowledgeable of its details since only a few of the Sanskrit logic texts had by then been translated into English.

This paper examines the question of Indian influence and tries to estimate what aspects of Indian logic are likely to have played a role in the ideas of George Boole. Although scholars are agreed that Indian logic had reached full elaboration, it was expressed in a special technical language that is not easily converted into the modern symbolic form. Boole most definitely was aware of the general scope of Indian logic and known that its focus was the cognition underlying the logical operation and this is something he aimed in his own work. He was trying to mathematize the role of the cognition and he believed that algebra would be effective for this purpose. To the extent he was attempting to go beyond what he knew of Indian logic, he thought he could do so by using mathematics.

**Boole's algebra**
Boole's starting point was algebra with variables like x and y, and algebraic operations such as addition and multiplication. He wished to show that algebra had in it the potential to extend the applicability of logic as well as the capacity to handle an arbitrary number of propositions. This is how he put it in his *Laws of Thought* [1]:

> There is not only a close analogy between the operations of the mind in general reasoning and its operations in the particular science of Algebra, but there is to a considerable extent an exact agreement in the laws by which the two classes of operations are conducted. Of course the laws must in both cases be determined independently; any formal agreement between them can only be established `a posteriori by actual comparison. To borrow the notation of the science of Number, and then assume that in its new application the laws by which its use is governed will remain unchanged, would be mere hypothesis. There exist, indeed, certain general principles founded in the very nature of language, by which the use of symbols, which are but the elements of scientific language, is determined. [Section 6 of Chapter 1]

Boole's algebra is about classes. He says: "That the business of Logic is with the relations of classes, and with the modes in which the mind contemplates those relations." [9] He represents the universe of conceivable objects by 1 or unity. Given the objects which are Xs, he calls the class by the same symbol X, and he means by the variable x "an elective symbol, which represents the *mental operation of selecting* [my emphasis] from that group all the Xs which it contains, or of fixing the attention upon the Xs to the exclusion of all which are not Xs" [9].

In my view, this emphasis on the "mental operation of selection" is the key to his scheme for it enlarges the setting to a much bigger system than the "universe of discourse" (a concept generally attributed to De Morgan but used as a phrase for the first time in *The Laws of Thought*).

Given two classes X and Y, Boole wrote:
- $x$ = the class X,
- $y$ = the class Y,
- $xy$ = the class each member of which is both X and Y, and so on.

Since selecting objects from the same class leaves the class unchanged, one can write:

$$xx = x^2 = x$$

In like manner he took

| | |
|---|---|
| $1 - x$ = the class not-X | (1) |
| $1 - y$ = the class not-Y | (2) |
| $x(1 - y)$ = the class whose members are Xs but not-Ys | (3) |
| $(1 - x)(1 - y)$ = the class whose members are neither Xs nor Ys | (4) |

Furthermore, from consideration of the nature of the mental operation involved, he showed that the following laws are satisfied:

$$x(y + z) = xy + xz \qquad (5)$$
$$xy = yx \qquad (6)$$

$$x^n = x \tag{7}$$

From the first of these it is seen that elective symbols are distributive in their operation; from the second that they are commutative. The third he termed the *index law*, the heart of his system, which he believed deals with the election (or choosing) that constitutes the process of logical inference.

He concluded: "The truth of these laws does not at all depend upon the nature, or the number, or the mutual relations, of the individuals included in the different classes. There may be but one individual in a class, or there may be a thousand. There may be individuals common to different classes, or the classes may be mutually exclusive. All elective symbols are distributive, and commutative, and all elective symbols satisfy the law expressed by (7)."[9]

Given n classes x, y, z, … the universe can be partitioned into $2^n$ regions where the classes come together in different ways.

With a single class, you have only two sets the class X and its complement, the class "not X" which was represented by Boole as (1-x).

$$1 = x + (1 - x) \tag{8}$$

For two classes, X and Y, we have the situation in equation (9) or, equivalently, in Figure 1:

$$1 = xy + x(1 - y) + (1 - x)y + (1 - x)(1 - y) \tag{9}$$

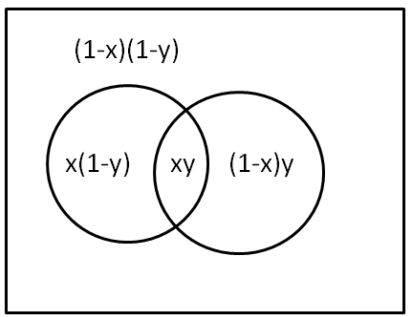

Figure 1. Four exclusive sets generated by classes X and Y (eqn. 9)

For three classes, X,Y, and Z, the situation is given by equation (10):

$$1 = xyz + xy(1-z) + x(1-y)z + (1-x)yz + x(1-y)(1-z) + (1-x)y(1-z) + (1-x)(1-y)z + (1-x)(1-y)(1-x) \qquad (10)$$

This is shown in Figure 2 below, where the region 1 is xyz; region 2 is xy(1-z) and so on.

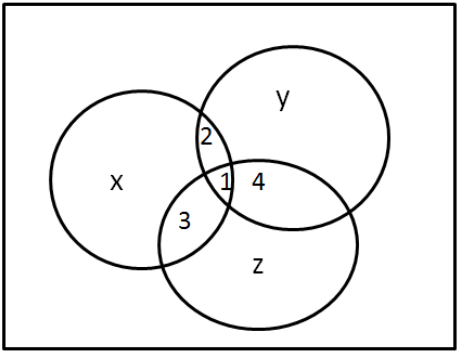

Figure 2. Division of the universe into 3 classes

Boole's system does not constitute what we know now as Boolean algebra, being different in fundamental ways. Boole's algebra X+Y cannot be interpreted by set union because $(X+Y)^2$ is not equal to (X+Y) as required by the condition (7):

$$(x+y)^2 = x^2 + y^2 + 2xy = x + y + 2xy$$

This is equal to (x+y) only under the restrictive condition that x and y are mutually exclusive.

Likewise, (x-y) is not a proper set since

$$(x+y)^2 = x^2 + y^2 - 2xy = x + y - 2xy$$

This is equal to (x-y) only under the condition that y is a subset of x, or xy =y. These two restrictive conditions are not consistent so Boole's algebra cannot be used in an effective manner for classes except to define subclasses as in (8) though (10).

What we now know as Boolean algebra is different from Boole's algebra, and why it works to the extent it does including division by 0 is now well understood [10]; Boolean algebra as we know was developed by Boole's successors (e.g. [11],[12]).

Boole speaks to the deeper foundations of logic in the concluding chapter of *The Laws of Thought*:

> [A]n evidence that the particular principle or formula in question is founded upon some general law or laws of the mind, and an illustration of the doctrine that the perception of such general truths is not derived from an induction from many instances, but is involved in the clear apprehension of a single instance. .. As in the pure abstractions of Geometry, so in the domain of Logic it is seen, that the empire of Truth is, in a certain sense, larger than that of Imagination. And as there are many special departments of knowledge which can only be completely surveyed from an external point, so the theory of the intellectual processes, as applied only to finite objects, seems to involve the recognition of a sphere of thought from which all limits are withdrawn.[1]

By this he means that a study of logic is likely to bring one closer to the domain of the mind that is informed by a deeper "infinite" truth. This is where intuitions of number and consequently of algebra were to be the bedrock of higher structures. This is why he was willing to use variables that were not immediately meaningful, for he believed they had the capacity to go beyond standard syllogisms.

**Indian logic: Nyāya and Navya Nyāya**
Let's now talk briefly of Indian logic (Nyāya), which has had a long history [13][14] that goes back to about 500 BCE. The stated goal in Nyāya is to state essential nature (*svarūpa*) that distinguishes the object from others. The fallacies of definition are that it is too broad, too narrow, or just impossible. There is also an old tradition that Greek and Indian logics are related [15] and that Kallisthenes, who was in Alexander's party, took logic texts from India and the beginning of the Greek tradition of logic must be seen in this material, but this is an issue that doesn't concern us here.

In Indian logic, minds are not empty slates; the very constitution of the mind provides some knowledge of the nature of the world. The four pramāṇas through which correct knowledge is acquired are perception (*pratyakṣa*), inference (*anumāna*), analogy (*upamāna*), and testimony (*śabda*).

Navya-Nyāya is a medieval elaboration of Indian logic [16][17]. It was founded by Udayana (c. 1050 CE), and further developed by Gaṅgeśa (c. 1200 CE), reaching its culmination in the works of Raghunātha (c. 1500 CE), Jagadīśa (c. 1600 CE) and Gadādhara (c. 1650 CE). The school developed a highly technical language. Its most famous text is Gaṅgeśa's *Tattvacintāmaṇi* ("Thought-Jewel of Reality") that deals with questions spanning logic, set theory, and epistemology.

Navya-Nyāya is concerned with describing the cognition of concern to the logician and its expression in language [18]. The cognition is supposed to be based on three facts: (i) two acts cannot be simultaneous; (ii) if introspection is to be possible, then the introspective act must follow immediately upon the object act, that is must not be more than one jump behind; (iii) the succeeding act constantly chases the preceding act. In other words, there is an attempt to analyze the very act of logical analysis.

The Nyāya technical language is not as elegant as a pure symbolic language and to process its claims is tedious [19][20]. In principle, this language can be converted into a symbolic form. The syntax of the language consists of relational abstract expressions, different kinds of term expressions —primitive, relational, abstract, and negative— and a negation particle.

A property with an empty domain was taken to be fictitious or unreal and non-negatable. Negation was considered a valid operation only on real properties. This could be considered to generate a three-valued table. If P, N, and U represent "positive", "negative", and "unnegatable", then we have the truth table [21]:

| w | not-w |
|---|---|
| P | N |
| N | P |
| U | U |

Knowledge was taken to be analyzed into three kinds of epistemological entities in their interrelations: "qualifier" (*prakāra*); "qualificand", or that which must be qualified (*viśeṣya*); and "relatedness" (*saṃsarga*). For each of these there was a corresponding abstract entity. Various relations were introduced, such as direct and indirect temporal relations, *paryāpti* relation (in which a property resides in sets rather than in individual members of those sets), *svarūpa* relation (which holds, for example, between an absence and its locus), and relation between the cognition of a knowledge and its object.

The concept of "limiterness" was used to put limits on time, property, and relations. The notion of negation was developed beyond specifying it with references to its limiting counterpositive, limiting relation, and limiting locus. The power of the technical language becomes clear when it is noted that questions such as the following were asked: Is one to recognize, as a significant negation, the absence of a thing A so that the limiter of the counterpositive A is not A-ness but B-ness? Gaṅgeśa said that the answer to these three questions was in the negative but he thought that the absence of an absence itself could lead to a new property suggesting consideration of higher abstractions.

According to Chakrabarti [19], Navya Nyāya anticipated several aspects of modern set theory. He explains:

> As an example Udayana says that there can be no universal of which every universal is a member; for if we had any such universal, then, by hypothesis, we have got a given totality of all universals that exist and all of them belong to this big universal. But this universal is itself a universal and hence (since it cannot be a member of itself, because in Udayana's view no universal can be a member of itself) this universal too along with other universals must belong to a bigger universal and so on ad infinitum. What Udayana says here has interesting analogues in modern set theory in which it is held that a set of all sets (i.e., a set to which every set belongs) does not exist. [19]

An overall summary is provided by Staal [22], "The representation of logical structures by means of Sanskrit expressions in Indian logic constitutes a formalization which is similar to the formalization adopted by Western symbolic logic. The various technical terms, the formation of compounds, the morphological means of expression (e.g. suffixes and case endings) and the syntactical means of expression (e.g. appositional clauses) in the technical Sanskrit of Navya-Nyāya are analogous to the terms, the formulas and the rules of modern Western logic."

In passing, we must add that Nyāya complements other approaches to reality through consideration of physical entities and the interaction of the observer and the observed [23][24][25], that have much affinity with quantum logics [26][27]. In the Indian physics tradition of Vaiśeṣika, the interaction between matter and mind was viewed through the idea of samavāya [28], which opens up several points of conceptual overlap with post-classical conceptions of reality (e.g. [23][29]). This is relevant here because of Boole's claim quoted above for he believed that intellectual processes constitute a "sphere of thought from which all limits are withdrawn." This willingness to confront "infinity" shows up in the details of Boole's algebra as described next.

**Boole's Interpretation Procedure**

Boole considered his algebraic methods for doing logic to be sound so long as he could interpret the end formula correctly. However, the expressions in the derivation had terms that could not have the usual meaning associated with the variables. For example, what does the equation xw = y mean about the class w? Boole solves this equation for w, obtaining w=y/x, and then expands it out.

In order to show how it was done, he took the algebraic function f(x) to be given by what he called the Expansion Theorem [30]:

$$f(x) = f(1) x + f(0) (1-x) \qquad (11)$$

He claimed this was an identity by taking f(x) = a x + b (1-x), where a and b can be computed by putting x=0 and x=1, respectively by what he called the Elimination Theorem [31]. Using x' as abbreviation for (1-x), he now wrote

$$f(x,y) = f(1,1) xy + f(1,0) xy' + f(0,1) x'y + f(0,0) x'y' \tag{12}$$

Thus if f(x,y) = y/x, he wrote this out without worrying about division by zero:

$$y/x = 1. xy + 0. xy' + 0/0 \, x y + 1/0 \, x'y \tag{13}$$

The term 0/0 contributes an indeterminate component to y/x as shown for v can be any subset of x'y' and the 1/0 term is to be solved separately by the side-condition that x'y = 0.

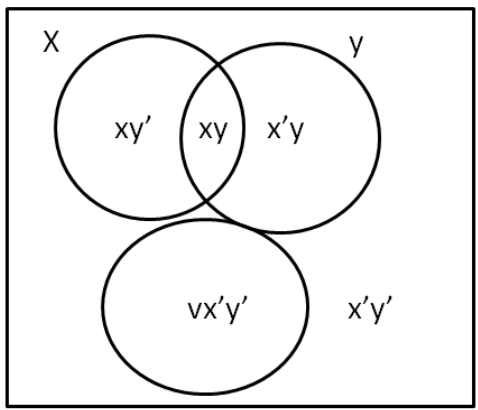

Figure 3. Division of the universe into 3 classes, where v is indeterminate

This keeps open the possibility that there are unknown other sets that can be part of the inference. The fact that an additional set outside of x and y could be associated with the problem must have been a point of attraction to Boole, although it does not provide any benefits and muddies up the analysis.

Perhaps the logic of going beyond x and y and consideration of the indeterminate component paralleled the idea of catușkoți, which has long been part of Indian logic. It has four components: P stands for any proposition and Not-P stands for its complement; both P and Not-P represents the usual universe; but the fourth part here is neither P not Not-P. The fourth part represents going beyond the domain of P and Not-P.

**Discussion**

This article suggests explanations for George Boole's development of his algebra for logic and the origins of its inadequacies. We have suggested broad parallels between his ideas and certain aspects of Indian logic that support his wife Mary Boole's assertion that he was deeply influenced by Indian logic. We argue that Boole's focus was more than a framework for propositions and that he was trying to mathematize cognitions as in the tradition of Indian logic and this is consistent with his own assertion that laws of thought should not be constrained by finitude. This may explain why he was happy to use operations in his algebra that allowed division by zero, which required further side-rules to eliminate infinities so that the final results were correct.